\documentclass[a4paper,conference]{IEEEtran}
\usepackage{cite}
\usepackage{amsmath,amssymb,amsfonts}
\usepackage{algorithmic}
\usepackage{graphicx}
\usepackage{textcomp}
\usepackage{color}
\usepackage{hyperref}
\usepackage{caption}
\usepackage{subcaption}
\usepackage{tikz}
\usepackage{pgfplots}

\begin{document}
\title{Facial Expression Recognition Using Residual Masking Network}

\author{
\IEEEauthorblockN{Luan Pham\IEEEauthorrefmark{2}, The Huynh Vu \IEEEauthorrefmark{2}, Tuan Anh Tran \IEEEauthorrefmark{1}\IEEEauthorrefmark{2}}
\IEEEauthorblockA{\IEEEauthorrefmark{2} Research Department-Cinnamon AI, Viet Nam}
        
\IEEEauthorblockA{\IEEEauthorrefmark{1} Faculty of Computer Science $\&$ Engineering, Ho Chi Minh City-University of Technology (HCMUT), Viet Nam} %\\ Corresponding author}

\IEEEauthorblockA{phamquiluan@gmail.com, vuthe\_huynh@yahoo.com, trtanh@hcmut.edu.vn}
}
\maketitle
%----------------------------------Abstract-------------------------------------------
\begin{abstract}
Automatic facial expression recognition (FER) has gained much attention due to its applications in human-computer interaction. Among the approaches to improve FER tasks, this paper focuses on deep architecture with the attention mechanism. We propose a novel Masking idea to boost the performance of CNN in facial expression task. It uses a segmentation network to refine feature maps, enabling the network to focus on relevant information to make correct decisions. In experiments, we combine the ubiquitous Deep Residual Network and Unet-like architecture to produce a Residual Masking Network. The proposed method holds state-of-the-art (SOTA) accuracy on the well-known FER2013 and private VEMO datasets. The source code is available at \url{https://github.com/phamquiluan/ResidualMaskingNetwork}.

\textit{Keywords: Facial Expression Recognition, Masking Idea, Residual Masking Network.}
\end{abstract}

\IEEEpeerreviewmaketitle
%-----------------------------------------------Introduction------------------------------------------
\section{Introduction}

Facial expression is one of the means of non-verbal communication, which accounts for a significant proportion of human interactions \cite{samadiani2019review,mehrabian2008communication}. It can be represented as discrete states (such as anger, disgust, fear, happiness) \cite{ekman1971constants} based on cross-culture studies \cite{matsumoto1992more}. Human emotions are sometimes mixed together in specific time and space conditions. However, ignoring the intricate emotions intentionally created by humans, the primary emotion is still widespread due to its intuitive definition. Like most other FER methods, our approach focuses on recognizing six facial emotional expressions (proposed by Ekman et al. \cite{ekman1971constants}) and the neutral state on the static image and ignoring the temporal relationships \cite{li2018deep}.

The FER task presents several challenges, especially in-the-wild settings due to the difference between inter-subject and intra-subject. For inter-subject variations, faces of individuals vary depending on different gender, ages, or ethnic groups. On the other hand, intra-subject changes include occlusions, illumination, and variations of head poses. Despite challenges, research in FER has drawn much attention, leading to several practical applications in human-computer interaction systems and data analytics \cite{li2018deep,martinez2016advances} such as sociable robotics, advertising, consumer behaviors or medical treatment.

In recent times, with the popularity of deep learning, especially convolutional neural networks, deep features can be automatically extracted and learned for a good facial expression recognition system \cite{li2018deep}. However, in the case of facial expressions, it is noteworthy to mention that much of the cues come from a few facial regions like eyes, mouth while other regions have little contribution to the output, e.g. hair, jawline. There are methods trying to focus on these important regions by using an  intermediate  feature named  facial  landmark  \cite{fan2018multi}. Facial landmarks detection can get  striking results in lab-controlled condition, but in a noisy environment, they usually not perform very well due to the variants in head poses, illumination, etc., see Figure \ref{fig:landmark_vs_attention}. 
Attention mechanism for the image classification problem in recent studies has been developed to increase the performance of the convolution neural network by focusing on tiny details \cite{woo2018cbam,park2018bam,wang2017residual}. Besides, in image segmentation problems, top-down bottom-up like architectures can keep useful information in pixel-level effectively \cite{ronneberger2015,li2018deepunet}. From these points, we propose a novel Masking Idea. This idea uses a Unet-based localization network to refine input feature maps and generates output feature maps that contain attention to some areas of the input feature maps. Those localization networks are called Masking Block.

Each Masking Block is a small variance of the Unet network \cite{ronneberger2015}, which enables the Residual Masking Network to focus on crucial spatial information and to make correct emotional expression classification. 

In this research, we provide three major contributions:

\begin{itemize}
    % \item Propose a novel deep architecture for emotion classification.
    \item Proposing a novel Masking Idea - an attention mechanism which can be embedded into convolutional neural network to boost the performance.

    \item Based on the advantage of Masking Idea to build a Residual Masking Network for facial expression recognition problem.

    \item Besides FER2013, a new dataset named VEMO is created in order to evaluate our network. 
\end{itemize}

Besides, the Masking Block used in this paper can be easily integrated into the existing networks. Our works are available online at Github \cite{luanresmaskingnet2020}.

The remaining parts of this paper are organized as follows: Section \ref{related_work} provides a brief review of previous related studies, Section \ref{proposed_method} describes the methods used for the network architecture, Section \ref{exerimental} includes a discussion about experimental results, and the paper is concluded in Section \ref{conclusion}.

%-----------------------------------------------Related work -----------------------------------------
\section{Related works} \label{related_work}

Even though the deep learning approaches recently show the efficiency in automatic facial expression recognition, traditional machine learning methods remain prevalent. These methods are still active in many cases and continue to be developed, such as \cite{Hong_2017}, \cite{Wang_2015}, \cite{Xiaohua_2016}, \cite{Saurav_2019}.
 
Traditional FER approaches have been using handcrafted features and experiments mostly on lab-controlled datasets. 
Features taken in places such as the eyes, nose, and mouth must be consistent across the image. 
Some well-known feature extractions can be listed as Local Binary Patterns (LBP) \cite{Xiaohua_2016}, LBP on Three Orthogonal Planes (LBP-TOP) \cite{Hong_2017}, Non-negative Matrix Factorization (NMF) \cite{Ali_2015}, Sparse Learning \cite{Deepak_2013}. 

In most of the traditional approaches, the first step in practice is to detect the position of the face, then, extract geometric features \cite{Deepak_2013}, appearance features \cite{SL_2012}, or both \cite{Byoung_2018} to generate specific vectors to the model. These methods are generally quite complex, requiring a lot of technical manipulation. The characteristic analysis becomes extremely challenging as the data is massive. These methods often face the problem in natural or noise environments where landmark detection is difficult.  

Combining traditional methods with deep learning can be an effective solution to this problem. A number of combined methods are proposed by \cite{connie2017facial} \cite{Deepak_2013} \cite{georgescu2019local}. As we can see in Table \ref{table:evaluation_FER2013}, their results are quite well. 
The main problem, however, is still the complexity of technical manipulation. Besides, those methods are usually designed optimally for a specific data set (or a particular target), which leads to low re-usability.

Leading to higher accuracy compared to traditional handcrafted features, recent static image FER has followed deep learning approaches. To prevent overfitting, lots of training images are required for deep learning-based networks. The introduction of more large scale datasets such as EmotioNet \cite{fabian2016emotionet}, AffectNet \cite{mollahosseini2017affectnet}, ExpW \cite{zhang2018facial}, FER2013 \cite{goodfellow2013challenges}, and the increasing computational power (such as GPUs, TPU) enabled more applications of these approaches.

Many CNN architectures were applied for FER to increase the expressiveness of feature representation. Yao et al. \cite{yao2016holonet} introduced the HoloNet by incorporating concatenated RELU (CRELU) \cite{shang2016understanding} and inception-residual blocks \cite{szegedy2016rethinking} to the existing Resnet structure \cite{he2016deep} to increase the network's depth and improve the multi-scale learning. In another design proposed by \cite{hu2017learning}, three supervised blocks were applied to raise the degree of supervision of the Resnet network \cite{he2016deep}.  

The Network ensemble is another strategy being applied for FER to improve its accuracy. Individual networks are ensembled by concatenating their features \cite{liu2016facial} or taking an average of their output predictions \cite{pons2017supervised}. For the ensembling to take effect, the network should have sufficient diversity by being trained on different training data, having varied architectures, parameters, size of filters, or several network layers. Hamester et al. \cite{hamester2015face} combined convolutional neural networks (CNNs) with a convolutional autoencoder (CAE) for architecture ensemble.

In traditional CNNs for the classification tasks, the loss is often applied to keep features of a different class apart. To increase the discrimination among different emotion expressions, Cai et al. \cite{cai2018island} proposed the island loss to decrease the variations among intra-class while expanding the difference among inter-class simultaneously.

\begin{figure}
	\includegraphics[width=0.45\textwidth]{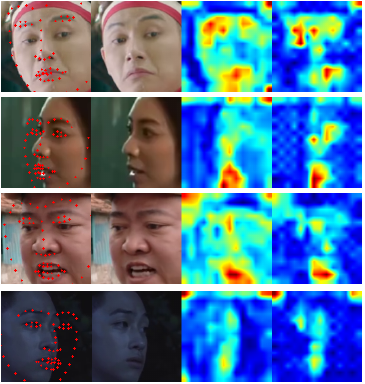}
	\centering
	\caption[center]{Example of landmark detections and features of Masking Block as follows: landmark detection, original image, feature map before the $3^{rd}$ Masking Block, feature map after the $3^{rd}$ Masking Block.}
	\label{fig:landmark_vs_attention}
\end{figure}

Summarizing state of the art in FER, We can draw the following key points as a basis for developing our network. Firstly, the accuracy of detecting essential facial areas contributes mainly to the improvement of the classification accuracy. Secondly, Facial expressions are determined based on the combination of some facial regions such as the eyes, nose, mouth \cite{fan2018multi}. Several traditional methods extracted these facial areas based on facial landmarks. However, the landmark detection worked well mostly on lab-controlled datasets but not in-the-wild datasets due to intra-subject variations such as occlusions, illumination, and variations of head poses (as shown on the first column of Figure \ref{fig:landmark_vs_attention}). Thirdly, For CNN-based FER approaches, the localization of facial areas can be observed through intermediate layers of CNN (the third and fourth column of Figure \ref{fig:landmark_vs_attention}). Also, attention mechanisms were often applied to CNN-based approaches to improve a network's concentration on relevant information and ignore unnecessary ones.

Wang et al \cite{wang2017residual} developed the Attention Module including a trunk branch and a mask branch with the argument that the trunk branch will perform the feature processing and mask branch will produce the same size mask that softly weights the output features of trunk branch. Their research results are very good and are proved by the empirical experiments. Analysis of this idea, we have two proposals:

\begin{itemize}
    \item It will be better if the output of the mask branch can score the importance of the output activation maps of the trunk branch.
    
    \item A deeper network might achieve a better result in localizing the important score of the feature maps, \cite{li2018deepunet}.
\end{itemize}
The masking idea argues that a localization network might help to refine the tensors by producing its importance weights, which supports the learning process to focus on what it deems necessary. The loss function would track the refinement. This intuition is the same as the way the U-net network receives a biological image and returns a segmentation mask.

\begin{table}

	\centering
% 	\caption{Residual Masking Network detailed configurations (RL: Residual layer, MB: Masking Block).}
    \caption{The detailed Residual Masking Network configuration (RL: Residual Layer, MB: Masking Block).}
	\label{table:resmasking_architecture}
	\begin{tabular}{l c r}
		\hline
		\textbf{Layer name} & 
		\textbf{Ouput size} &
		\textbf{Detail} \\
        \hline
		Conv1    & $64\times 112\times 112$ & $7\times 7$, stride 2 \\
		MaxPooling  & $64\times 56\times 56$  & $3\times 3$, stride 2\\
		Resmasking Block 1 & $64\times 56\times 56 $ & RL 1, MB 1 \\
		Resmasking Block 2 & $128\times 28\times 28$ & RL 2, MB 2 \\
		Resmasking Block 3 & $256\times 14\times 14$ & RL 3, MB 3 \\
		Resmasking Block 4 & $512\times 7\times 7$ & RL 4, MB 4 \\
		Average pooling & $512\times 1\times 1$ \\
		FC, Softmax & $7$ \\
		\hline
	\end{tabular}%
\end{table}%

%----------------------------------------------- Proposed method --------------------------------
\section{Proposed method} \label{proposed_method}

\begin{figure*}
	\includegraphics[width=\textwidth]{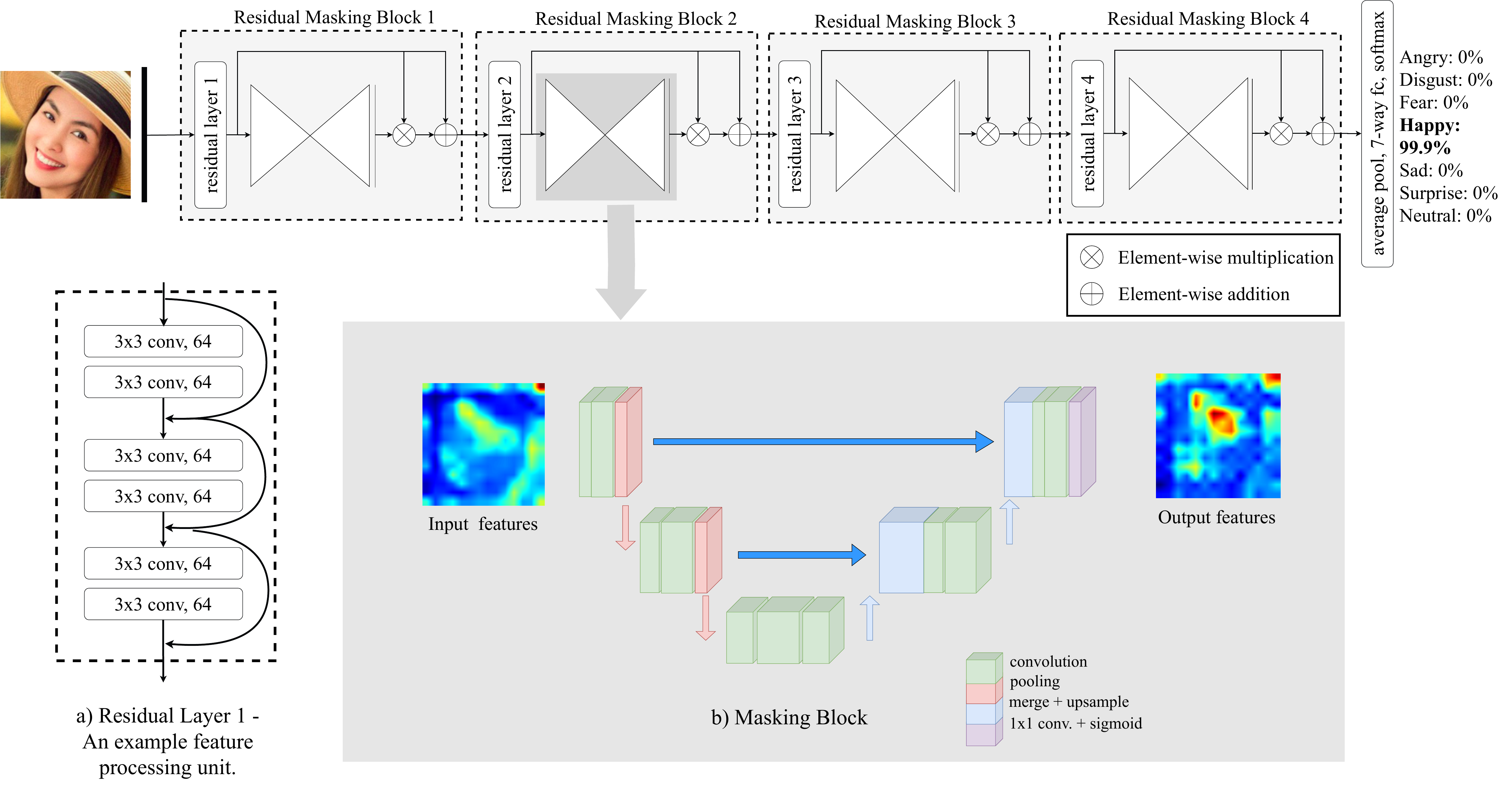}	
	\caption{The overview of Residual Masking Network.}
	\label{fig:residual_masking_network_overview}
\end{figure*}

\subsection{Overview}

The main flow of the proposed method is the Residual Masking Network illustrated in Figure \ref{fig:residual_masking_network_overview}.
This network contains four main Residual Masking Blocks (Resmasking Blocks). Each Residual Masking Block, which operates on different feature sizes, contains a Residual Layer and a Masking Block (see Table \ref{table:resmasking_architecture}). 

An input image of size $224\times224$ will go through the first $3\times3$ convolutional layer with stride $2$ before passing a $2\times2$ max-pooling layer, reducing its spatial size to $56\times56$. Next, the feature maps obtained after the previous pooling layer are transformed by the following four Residual Masking Blocks with generated features maps of four spatial sizes, including $56\times56$, $28\times 28$, $14\times 14$, and $7\times 7$.
The network ends with an average pooling layer and a 7-way fully-connected layer with softmax to produce outputs corresponding to seven facial expression states (6 emotions and one neutral state).

\subsection{Residual Masking Block}
In this research, we propose the Masking Block, which performs the scoring operation. Then, input feature maps of the Masking Block and its outputs are directly combined. We remove the trunk branch and re-use Resnet34 \cite{he2016deep} as the backbone.

We design Residual Masking Block containing a Residual Layer and a Masking Block, with the former being in charge of feature processing and the latter producing the weights for the corresponding feature maps, as in Figure \ref{fig:residual_masking_network_overview}.

Given an input feature map $F \in R^{C \times W \times H}$, firstly, $F$ will go through Residual Layer $R$ (Figure \ref{fig:residual_masking_network_overview}a) to produce the coarse feature map $F_R = R(F)$, $F_R \in R^{C' \times W' \times H'}$.
Then, a same size activation map $F_M$ having value in range $[0, 1]$ are calculated via Masking Block by the formular $F_M = M(F_R)$. Finally, the refined feature map - output of Residual Masking Block will be yielded by the Formula \ref{eq:remasking function}. This way, we assume that $F_M$ would be more convenient to score the element-wisely importance of the input feature map $F_R$ than the changed one \cite{wang2017residual}.

\begin{equation} \label{eq:remasking function}
F_N = F_R + F_R \otimes F_M,
\end{equation}

where $F_R$ is a transformed feature map of $F$ via the Residual Layer, $\otimes$ denotes the element-wise multiplication. We also use attention residual learning method proposed in \cite{wang2017residual} to prevent the Masking Block from removing good features.

Masking Block bases on the Unet structure proposed in \cite{ronneberger2015}, which is a famous structure to localize small medical objects. This block consists of one contracting path (encoder) and one expansive path (decoder) as shown in Figure \ref{fig:residual_masking_network_overview}b. 
The use of the Masking Block is the main different of our attention modules compared to others \cite{woo2018cbam,wang2017residual}. 
In addition, it is noteworthy that the Masking Block have varied number of pooling and upsampling layers depending on the feature spatial size of input residual unit.

 It should be noted that the Masking Block can play a role of activations in many segmentation architectures instead of only Unet-like experiment as in our method.
\subsection{Ensemble Method}
In competitive environment, it is difficult to avoid the impact of ensemble methods on booming accuracy. To demonstrate the ability to combine the Residual Masking Network with other CNNs, we use a simple no-weighted sum average ensemble to fuse the prediction results of $7$ different CNNs. The model searching is based on the terms of validation accuracy as in \cite{pramerdorfer2016facial}. The procedure of generating ensemble results is described at our Github \cite{luanresmaskingnet2020}.

%----------------------------------Experiment----------------------------------
\section{Experimental results} \label{exerimental}

\subsection{Dataset}
\begin{figure}[!tb]\label{fig:example_dataset}
\begin{subfigure}{0.46\textwidth}
	\includegraphics[width=\textwidth]{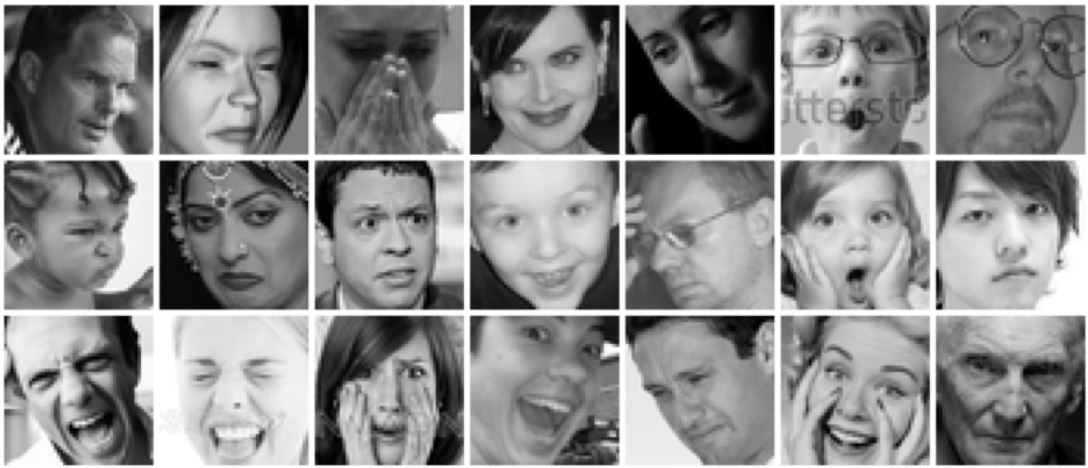}
	\centering
	\caption[center]{FER2013 dataset}
	\label{fig:FER2013}
   \end{subfigure}
   
   \begin{subfigure}[b]{0.46\textwidth}
	\includegraphics[width=\textwidth]{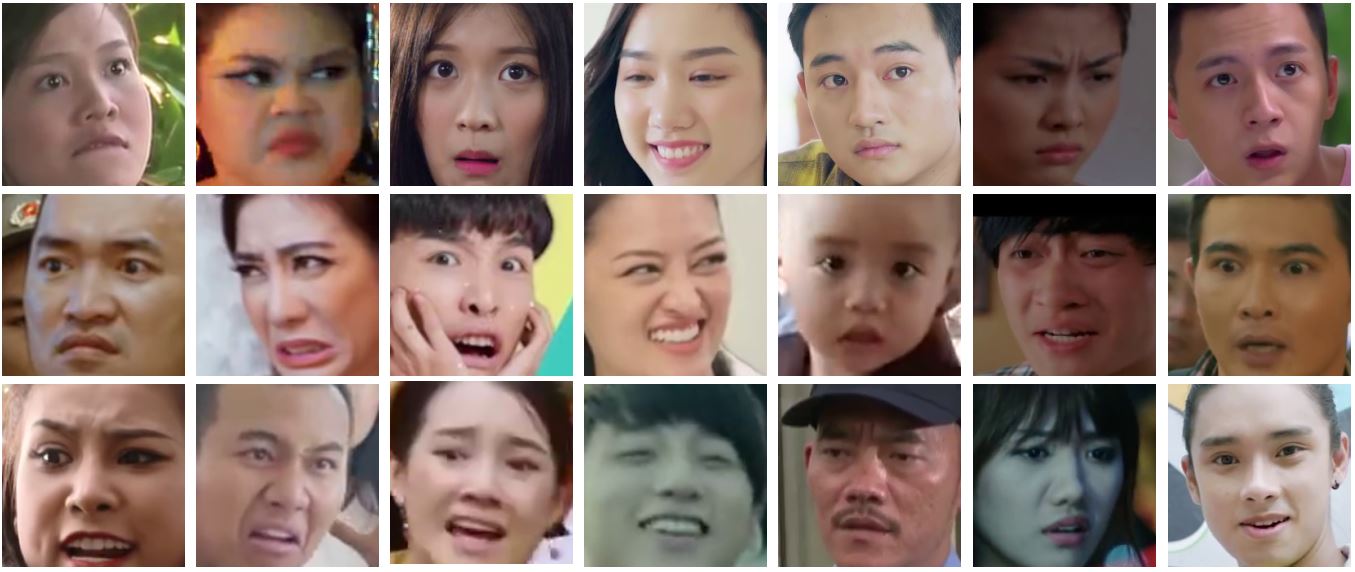}
	\centering
	\caption[center]{VEMO dataset}
	\label{fig:VEMO}
	\end{subfigure}
	\caption{Example images of two datasets.}
\end{figure}

To test the efficiency of the proposed method, experiments were conducted on one published dataset and one private dataset, which is going to be public in the near future. 
The first well-known dataset is FER2013 \cite{goodfellow2013challenges}, which was introduced during the ICML 2013 Challenges in Representation Learning. 
This dataset, as shown in Figure \ref{fig:FER2013}, contains a total of 35887 grey-scale (48x48) images. 
There is a total of 28709 images used for training images, 3859 for validation and 3589 for testing. 

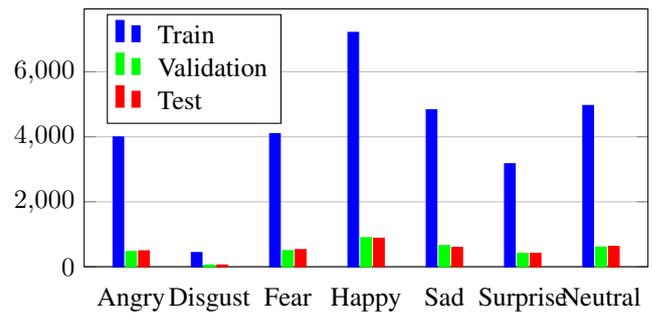
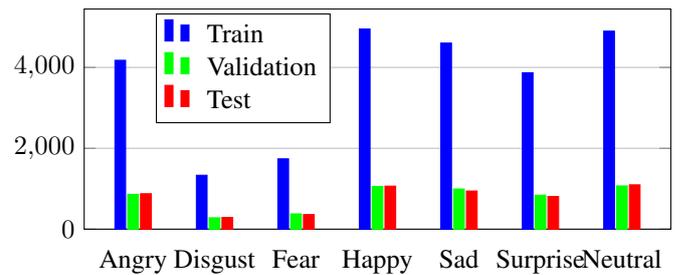
\begin{figure}[!tb]\label{fig:data_statistic}
\begin{subfigure}{0.46\textwidth}
\begin{tikzpicture}
    \begin{axis}[
        width  = 9cm,
        height = 5cm,
        major x tick style = transparent,
        ybar=2*\pgflinewidth,
        bar width=4pt,
        ymajorgrids = true,
        ylabel = { },
        symbolic x coords={Angry,Disgust,Fear,Happy,Sad,Surprise,Neutral},
        xtick = data,
        scaled y ticks = false,
        enlarge x limits=0.1,
        ymin=0,
        legend cell align=left,
        legend style={
                at={(0.35,0.98)},
                anchor=north east,
                column sep=1ex
        }
    ]
        \addplot[style={blue,fill=blue,mark=none}]
            coordinates {(Angry,3995) (Disgust,436) (Fear,4097) (Happy,7215) (Sad,4830) (Surprise,3171) (Neutral,4965)};

        \addplot[style={green,fill=green,mark=none}]
            coordinates {(Angry,467) (Disgust,56) (Fear,496) (Happy,895) (Sad,653) (Surprise,415) (Neutral,607)};

        \addplot[style={red,fill=red,mark=none}]
            coordinates {(Angry,491) (Disgust,55) (Fear,528) (Happy,879) (Sad,594) (Surprise,416) (Neutral,626)};

        \legend{Train,Validation,Test}
    \end{axis}
\end{tikzpicture}

	\caption[center]{FER2013 dataset}
	\label{fig:FER2013Statistic}
   \end{subfigure}

   \begin{subfigure}[b]{0.46\textwidth}

\begin{tikzpicture}
    \begin{axis}[
        width  = 9.3cm,
        height = 4.5cm,
        major x tick style = transparent,
        ybar=2*\pgflinewidth,
        bar width=4pt,
        ymajorgrids = true,
        ylabel = { },
        symbolic x coords={Angry,Disgust,Fear,Happy,Sad,Surprise,Neutral},
        xtick = data,
        scaled y ticks = false,
        enlarge x limits=0.1,
        ymin=0,
        legend cell align=left,
        legend style={
            at={(0.42,0.98)},
            anchor=north east,
            column sep=1ex
        }
    ]
        \addplot[style={blue,fill=blue,mark=none}]
            coordinates {(Angry,4176) (Disgust,1333) (Fear,1739) (Happy,4948) (Sad,4601) (Surprise,3867) (Neutral,4896)};

        \addplot[style={green,fill=green,mark=none}]
            coordinates {(Angry,860) (Disgust,281) (Fear,378) (Happy,1054) (Sad,993) (Surprise,839) (Neutral,1069)};

        \addplot[style={red,fill=red,mark=none}]
            coordinates {(Angry,876) (Disgust,289) (Fear,361) (Happy,1064) (Sad,943) (Surprise,807) (Neutral,1096)};

        \legend{Train,Validation,Test}
    \end{axis}
\end{tikzpicture}

	\caption[center]{VEMO dataset}
	\label{fig:VEMO_statistic}
	\end{subfigure}
	\caption{The statistics of training, validation, and testing set.}
% 	\caption{The statistics in the number of images of training, validation, and testing set across seven emotional expressions include Angry, Disgust, Fear, Happy, Sad, Surprise and Neutral.}
\end{figure}

Each image is collected by Google image search API and labeled as one of the seven categories, including anger, disgust, fear, happiness, sadness, surprise and neutral. 
This dataset is widely used for evaluating deep learning-based FER methods. 
However, the dataset contains several invalid samples (e.g. non-face images or images with faces cropped incorrectly) and the image distribution among emotion categories is not equal. 
As shown in Figure \ref{fig:FER2013Statistic}, there are more than 6000 images showing happiness (H) while the number of images containing disgust (D) is just approximately $500$.

The second dataset is Vietnam Emotion (VEMO2020). This dataset contains 36470 images (in multi-resolution) that are separated into two parts. The first part contains 6470 colored photos that are collected from Youtube, which are applied \cite{Paul_2001} to detect face region and then select five frames per second for each video; and from Google Image, Flickr (Vietnamese people only). 
The emotion of each image is chosen by the voting of a group including ten members (each member from $18$ to $23$ years old). 
The second part includes 30000 images that were labeled by the professional in emotion labeling \cite{mollahosseini2017affectnet}. The examples and the distribution among facial expression categories in training/validation/testing set of this dataset are shown in Figure \ref{fig:FER2013Statistic} and Figure \ref{fig:VEMO_statistic}.

\subsection{Experiment setup}

The original training images are scaled up to 224x224 and converted to RGB before the training process to adapt with ImageNet pretrained models. 
Besides,  training images are augmented to prevent overfitting. 
The augmentation methods include left-right flipping and rotating in the range of $[-30, 30]$. 

Each experiment last for maximum 50 epochs, and stop when validation accuracy is not improved more than 8 steps, given the batch size of 48 and the initial learning rate of 0.0001, scheduler reduces learning rate 10 times if validation accuracy not increased in two continuous epoches. 

The momentum is set to 0.9 and weight decay to 0.001. The evaluation metric for classification task is:
\begin{equation}
    Accuracy=\frac{TP+TN}{TP+TN+FP+FN}
\end{equation}
where TP: True Positive, TN: True Negative, FP: False Positive, FN: False Negative. 
Experiments from different networks are also conducted using the same setting environment such as hyperparameters, preprocessing, augmentation, as well as evaluation metrics.

\begin{figure}[!tb]
	\includegraphics[width=0.45\textwidth]{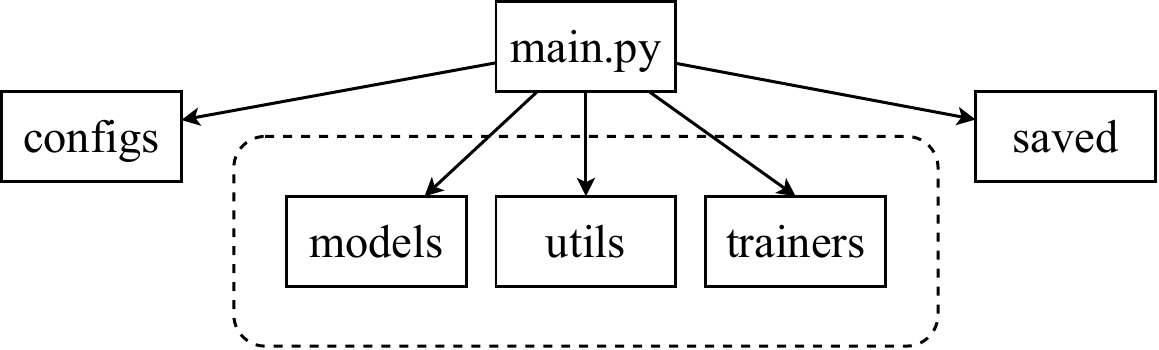}
	\centering
	\caption[center]{The framework structure of the experiment setup.}
	\label{fig:framework_structure}
\end{figure}

Experiments are conducted using Pytorch framework \cite{paszke2019pytorch} and Python language \cite{van1995python} on GTX 1080Ti. The framework structure could be found in Figure \ref{fig:framework_structure}. 
The detail of the experiments, reports, as well as the inference or testing code, is also presented in Github \cite{luanresmaskingnet2020}. On the other hand, a laptop CPU core I7-8750H, VGA GTX 1050Ti, Ram 16GB is used for testing processing time in the real application.
With this infrastructure, the proposed network can process 100 frames per second, each frame contains a single face.
With this result, we can guarantee the real-time application.
\subsection{Visual result explanation}
\begin{figure}[!tb]
	\includegraphics[width=0.45\textwidth]{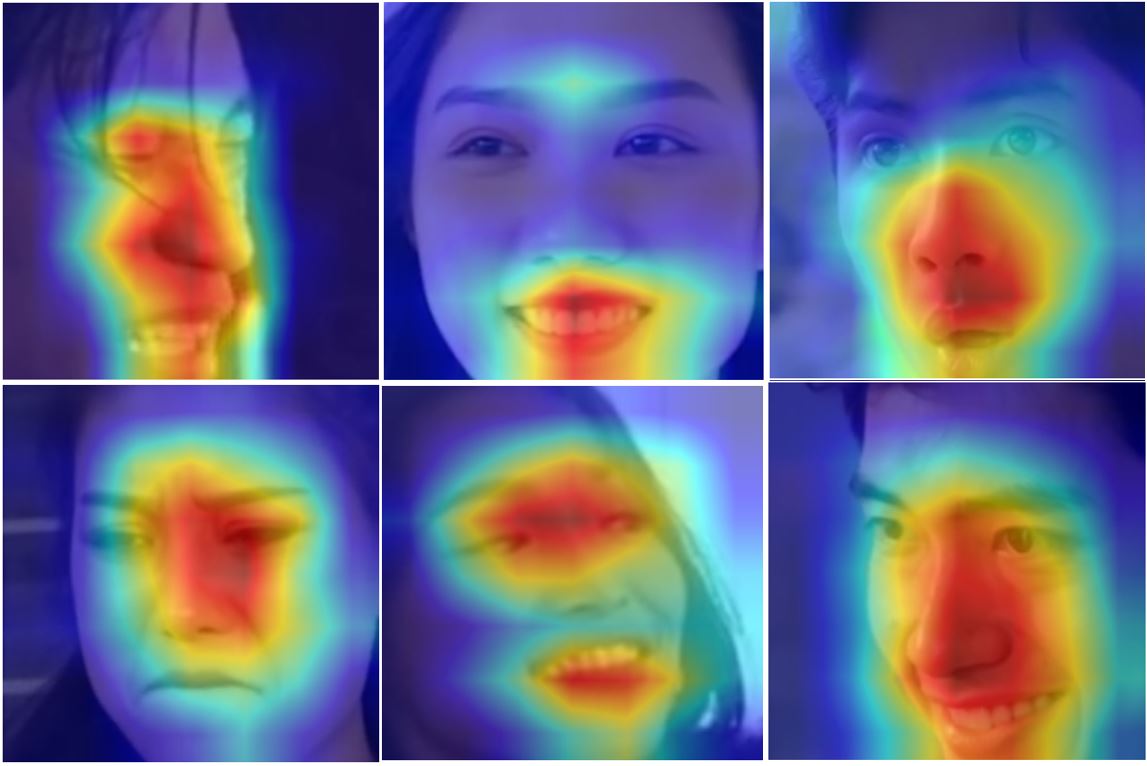}
	\centering
	\caption[center]{Grad-CAM visualization \cite{selvaraju2017grad} of some images in the testing sets.}
	\label{fig:Grad-CAM}
\end{figure}

To get a further understanding of which area in a face contributes to a classification result, we applied the Grad-CAM approach proposed by Selvaraju et al.
\cite{selvaraju2017grad} to obtain a visual explanation. This approach used the gradients of any class, routing to the final convolutional layers to generate a map showing which region in the images that predicts this class. Some visualizations of the Grad-CAM approach on test images of FER2013 are shown in Figure \ref{fig:Grad-CAM} with high activation area colored in red. The red area often positions near the mouth or the eyes, meaning that they are the essential facial areas that the network based on to make classification decisions.

\subsection{Evaluation and Analysis}
The proposed method has been tested on several aspects and produced positive results.

For the public dataset FER2013, firstly, we choose a number of well-known and powerful classification networks (such as Resnet151 \cite{he2016deep}, Densenet121 \cite{huang2017densely}, Cbam\_resnet50 \cite{woo2018cbam}, and Bam\_resnet50 \cite{park2018bam}) to evaluate our method (train/test in the same environment and dataset). Table \ref{table:result_with_popular_network} shows the comparison between them in the accuracy as well as the number of parameters. As presented in this table, although our Residual Masking Network has the highest number of parameters, it outperforms recent well-known deep learning-based classification networks.

\begin{table}[!t]
	\centering
	\caption{Performance evaluation of well-known classification networks and our method on FER2013}
	\label{table:result_with_popular_network}
	\begin{tabular}{l c r}
		\hline
		\textbf{Networks} & 
		\textbf{Parameters x $10^6$} & 
		\textbf{Accuracy (\%)} \\
        \hline
		VGG19 \cite{simonyan2014very}  & 139.5 & 70.80  \\
		Eficientnet\_b2b \cite{tan2019efficientnet}  & 7.7 & 70.80  \\
		Googlenet \cite{simonyan2014very} & 5.6  & 71.97  \\
		Resnet34 \cite{he2016deep}  & 21.2 & 72.41 \\
% 		Resnet50 \cite{he2016deep}  & 25.0 & 72.22  \\ 
		ResAttNet56 \cite{wang2017residual}  & 29.0 & 72.63  \\ 
		Inception\_v3 \cite{szegedy2016rethinking} & 25.1  & 72.72  \\
% 		Resnet18 \cite{he2016deep} & 11.2  & 72.9  \\
		Bam\_resnet50 \cite{park2018bam} & 23.8  & 73.14  \\
		Densenet121 \cite{huang2017densely} & 6.9  & 73.16  \\
		Resnet152 \cite{he2016deep} & 58.1 & 73.22  \\
		Cbam\_resnet50 \cite{woo2018cbam} & 28.5  & 73.39  \\
		
		\textbf{Our ResMaskingNet} &  \textbf{142.9}  & \textbf{74.14}  \\
		\hline
	\end{tabular}%
\end{table}%

Besides, the detailed evaluation with the SOTA networks on the FER2013 (ensemble /without ensemble mode) is presented in Table \ref{table:evaluation_FER2013}.   
With the ensemble mode, our Residual Masking Network ensembled with 6 CNNs obtained the highest result of 76.82\%, outperformed all ensemble-based methods on FER2013 by 1\%. On the other hand, in non-ensemble mode, our Resmaking network obtained the highest result of 74.14\%, more top than the second-highest result (73.14\%) of the CNN-SIFT \cite{connie2017facial}. It should be noted that the CNN-SIFT does not follow the single model; it aggregates features from both deep learning and SIFT algorithm.

For the VEMO dataset, we conducted four experiments to compare the performance of our Residual Masking Network with three other classification networks as following: Resnet18, Resnet34, and ResAttNet56 (see Table \ref{table:VEMO_evaluation}).

\begin{table}[!b]
	\centering
	\caption{Performance evaluation (Accuracy) of reported methods without ensemble (WE) and ensemble (E) on FER2013.}
	\label{table:evaluation_FER2013}
	\begin{tabular}{l c r}
		\hline
		\textbf{Networks} & \textbf{E (\%)} & \textbf{WE (\%)} \\
        \hline
		Human Accuracy \cite{goodfellow2013challenges}  & - & 65 $\pm 5$  \\
		DNN\_L\_Reg \cite{mollahosseini2016going}  & - & 66.4  \\
		Deep-Emotion \cite{minaee2019deep}   & - & 70.02  \\
		DL-SVM \cite{tang2013deep}    & - & 71.16  \\
		CNN-SIFT \cite{connie2017facial}    &- & 73.4  \\
		Ensemble N\&A \cite{kim2016fusing}  & 73.31  & -\\
		Ensemble MLCNNs \cite{nguyen2019facialensemble}  & 74.09  & -\\
		CNNs and BOVW $+$ local SVM \cite{georgescu2019local}  & 75.42 & - \\
		Ensemble 8 CNNs \cite{pramerdorfer2016facial}  & 75.2 &-  \\
		\textbf{Our ResMaskingNet} & \textbf{76.82}& \textbf{74.14}\\
		\hline
	\end{tabular}%
\end{table}%

\begin{table}[!ht]
	\centering
	\caption{Performance evaluation of three classification networks and our Residual Masking Network on the VEMO dataset.}
% 	\caption{Performance evaluation of three classification networks and our method on the VEMO dataset.}
	\label{table:VEMO_evaluation}
	\begin{tabular}{l r}
		\hline
		\textbf{Networks} & 
		\textbf{Accuracy (\%)} \\
        \hline
        ResAttNet56 \cite{wang2017residual} & 60.82  \\
		Resnet18 & 63.94  \\
		Resnet34 & 64.84  \\
	 \textbf{Our ResMaskingNet}  &  \textbf{65.94}  \\
		\hline
	\end{tabular}%
\end{table}%

\begin{figure}[!tb]
\begin{subfigure}{0.46\textwidth}
    \includegraphics[width=\textwidth]{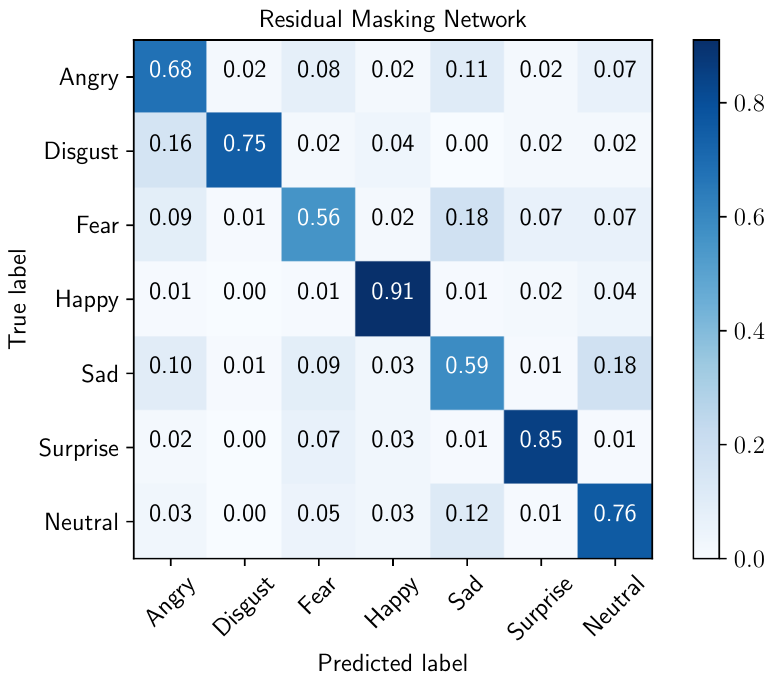}
	\centering
		\label{fig:confusion_matrix_FER2013}
	\caption[center]{FER2013 dataset}
   \end{subfigure}
   
   \begin{subfigure}[b]{0.46\textwidth}
	\includegraphics[width=\textwidth]{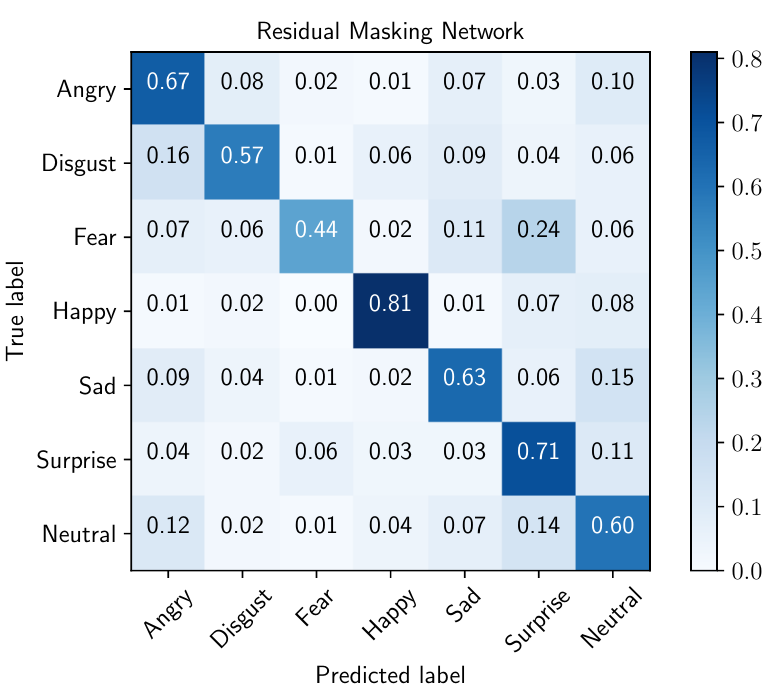}
	\centering
	\caption[center]{VEMO dataset}
	\label{fig:confusion_matrix_VEMO}
	\end{subfigure}
	\caption{Quantitative results in form of confusion matrix on two testing sets}
	\label{fig:confusion_matrix}
\end{figure}

\begin{figure}[!tb]
	\includegraphics[width=0.45\textwidth]{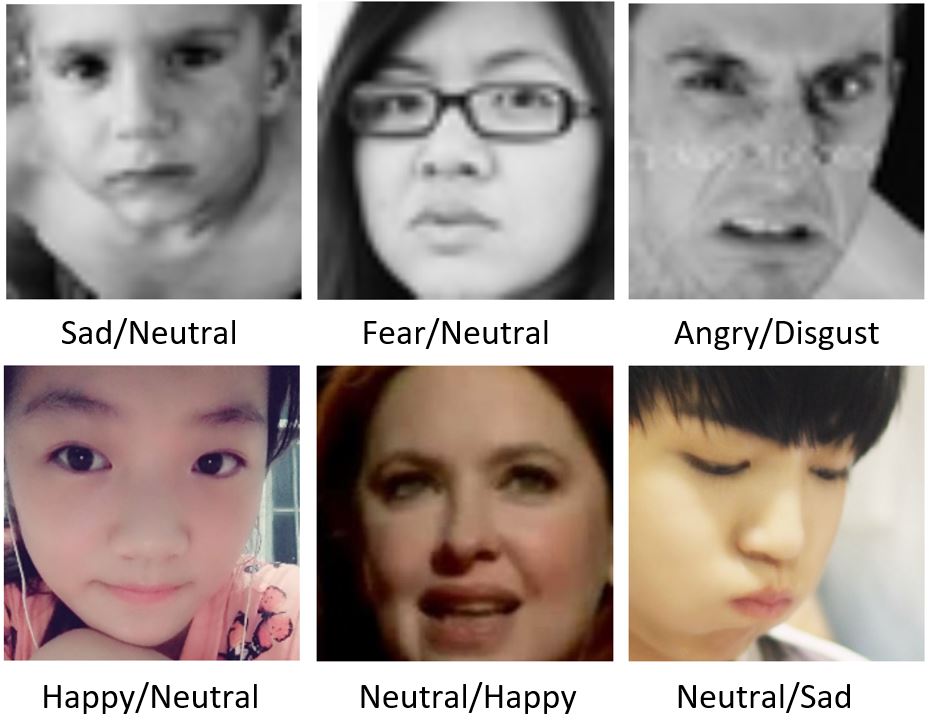}
	\centering
	\caption[center]{Examples of wrong emotion recognition on FER2013 and VEMO dataset (Ground truth / Predict)}
	\label{fig:Error_analysis}
\end{figure}

Besides the accuracy comparison, the Residual Masking Network's performance is also evaluated using confusion maxtrix as shown in Figure \ref{fig:confusion_matrix} where each row of the matrix represents instances of a predicted class while each column shows instances of the true label. The matrix values shows that the network performs well on almost emotions even though the FER2013 or VEMO training dataset is unbalanced across emotions. The Happy and Surprise emotion obtained the high scores of 0.91/0.81 and 0.85/0.71 while the performance of Fear and Sad are just 0.56/0.44 and 0.59/0.63, which are the lowest scores among emotions scores. Example of the detection errors can be found in Figure \ref{fig:Error_analysis}. 

We can see the ability of Masking Block in boosting the accuracy of the Resnet 34 model on both FER2013 and VEMO datasets. As shown in Figure \ref{fig:landmark_vs_attention}, where the activations before (column 3) and after (column 4) the $3^{rd}$ Masking block are visualized, the heat map seems to prefer the eyes, nose, and mouth to other facial regions. 

These results also appear to be in line with reality for a person's emotional recognition, it is a quite subjective process \cite{barsoum2016training}. Most of the wrong predictions come from wrongly labeling or the unclear emotion of the subjects, see Figure \ref{fig:Error_analysis}. About the wrongly labeling,  Barsoum et al \cite{barsoum2016training} tried to fix it and they also produce the FER+ dataset which is less noisy than the original one. 
A research result presented in \cite{goodfellow2013challenges} shows that the people's ability to recognize other human's emotions is just above the average. 
As presented in Table \ref{table:evaluation_FER2013}, the Human accuracy (accuracy given by a Human estimation) is only $65$\% with an error of $5\%$. The most recognizable emotions are happy and surprise, whereas the complex emotions like sadness or fear are difficult to distinguish.
Despite not playing a decisive role, data imbalance makes it a bit difficult for emotional recognition algorithms. 

Obviously, rare emotions like fear or disgust are harder to identify with either humans or machines. Emotion is still a challenging topic when people themselves are still very much confused about their feeling. This is also an exciting feature of this research direction.

%------------------------------Conclusion--------------------------------------
\section{Conclusion} \label{conclusion} 
This paper put forward a system for facial expression recognition,
in which the main contribution is a novel Masking Idea that is implemented in the Residual Masking Network. This  Residual Masking Network contains several Masking Blocks which are applied across Residual Layers to improve the network's attention ability on important information.  Experimental results showed that the proposed methods possess higher accuracy than the well-known classiﬁcation systems as well as the current state-of-the-art reported results on the FER2013 dataset. The focus on the future improvement of the proposed method is checking out the model generalization by evaluating it on the largest classification dataset, the ImageNet dataset. Furthermore, different network parameters, as well as model parameters reduction, will be explored to improve network performance across vision tasks such as classification and detection.
We have a goal of building a complete system and conducting testing on an open rehearsal environment.

\section*{Acknowledgment}
I would like to thank Mr. Pham Van Linh and Dr. Nguyen Trung Kien for his support in our research. Also, thank Ms. Pham Huong Linh for her management support.

% Generated by IEEEtran.bst, version: 1.14 (2015/08/26)

\end{document}